# Rotation-Invariant Point Convolution With Multiple Equivariant Alignments.

Hugues Thomas
University of Toronto
hugues.thomas@robotics.utias.utoronto.ca

## Abstract

*Recent attempts at introducing rotation invariance or equivariance in 3D deep learning approaches have shown promising results, but these methods still struggle to reach the performances of standard 3D neural networks. In this work we study the relation between equivariance and invariance in 3D point convolutions. We show that using rotation-equivariant alignments, it is possible to make any convolutional layer rotation-invariant. Furthermore, we improve this simple alignment procedure by using the alignment themselves as features in the convolution, and by combining multiple alignments together. With this core layer, we design rotation-invariant architectures which improve state-of-the-art results in both object classification and semantic segmentation and reduces the gap between rotation-invariant and standard 3D deep learning approaches.*

## 1. Introduction

The development of new 3D deep learning methods has been thriving in the past years. In particular, it is now possible to use convolutions on 3D point clouds, despite their sparse and unordered nature. Point convolution networks have reached unprecedented performances on tasks like semantic segmentation, object detection, or even point cloud registration. This success is allowed by the ability of such networks to remain invariant to point permutation. But point permutation is not the only invariance inherent to 3D data. Most of the time, 3D objects are seen under arbitrary rotations, at least along one axis. Rotation invariance is therefore central in point cloud processing. Many real scenes are naturally oriented by gravity, and are only free to rotate along the $z$ axis, but some objects can appear with an arbitrary rotation. For example, we can think of objects that fell on the ground. These cases are not the most common, but are the one that could cause failures in real applications.

To become robust to varying rotations, standard 3D point networks use data augmentation procedures, which are not sufficient to ensure generalizability against unseen rotations. A better approach is to design rotation-invariant methods, i.e. outputting the same result when their input is rotated by any given rotation. However, to cope with random rotations of objects inside a complete scene, these methods need to be more than rotation-invariant, they should be locally invariant by rotation, i.e. outputting the same result when local parts of their input are rotated by any given rotation.

Strict local rotation invariance is practically impossible to achieve as rotating a part also modifies the overall input shape. But it is possible to come closer to it, by using local operation that are invariant by rotation.

Recently, various works have proposed 3D deep learning methods that are invariant under global [6, 19, 25, 32, 33] and local [4, 15, 36, 37] rotations. Although these methods help the network ability to generalize to unseen rotations, they are still far from reaching the performances of standard deep learning methods in aligned scenarios. The gap is even larger when we consider only locally invariant methods.

In this works, we present a new rotation-invariant convolution operator to process point clouds. We use this operator to build deep neural network architectures that are invariant under local rotations. The following paragraphs present our contributions where key concepts are highlighted.

At the center of our work is the intrinsic link between rotation invariance and rotation equivariance, that we use to design our rotation-invariant convolution. As illustrated in Figure 1, **a convolution, like any mapping function, can be transformed into a rotation-invariant mapping, by aligning its input with the result of a rotation-equivariant mapping**, which is called Local Reference Frame (LRF) as in [38]. Furthermore the resulting rotation-invariant features can be transformed into a new LRF and realigned to be rotation-equivariant. This gives us the tools to pass both invariant features and equivariant LRFs from layer to layer in a deep neural network.

To be able to capture more accurate information, we propose two different upgrades for this aligned convolution. First, **we use the LRFs of the convolution neighbors as additional input features** for the convolution (See Figure 2). These LRFs are processed by a Multi-Layer Perceptron (MLP), to be resized to the right dimension, and concate-

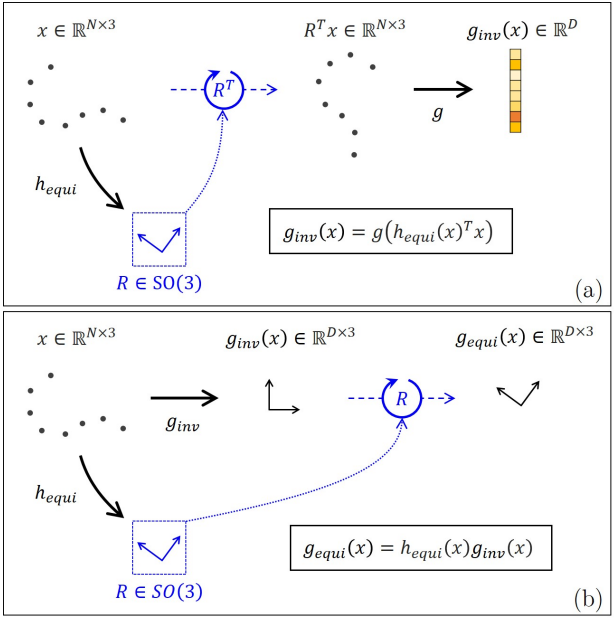

Figure 1. Illustration of the link between rotation invariance and rotation equivariance. (a) A rotation-invariant mapping $g_{inv}$ can be constructed from any mapping $g$ with a equivariant alignment $h_{equi}$. (b) A rotation-equivariant mapping $g_{equi}$ can be constructed from any rotation-invariant mapping $g_{inv}$ with a equivariant alignment $h_{equi}$.

**Definition 1** (Rotation invariance). *Let $N, D \in \mathbb{N}^+$ be our space dimensions. A mapping $g : \mathbb{R}^{3 \times N} \mapsto \mathbb{R}^D$ is invariant by rotation if $\forall x \in \mathbb{R}^{3 \times N}$ and $\forall R \in \mathrm{SO}(3)$,*

$$g(Rx) = g(x) \tag{1}$$

**Definition 2** (Rotation equivariance). *Let $N, D \in \mathbb{N}^+$ be our space dimensions. A mapping $g : \mathbb{R}^{3 \times N} \mapsto \mathbb{R}^{3 \times D}$ is equivariant by rotation if $\forall x \in \mathbb{R}^{3 \times N}$ and $\forall R \in \mathrm{SO}(3)$,*

$$g(Rx) = Rg(x) \tag{2}$$

Note that we let the output space be general, so these definitions still stand for mappings that have a pointwise output in $\mathbb{R}^{D \times N}$ or, using the matrix representation of rotations, in $\mathrm{SO}(3)^N \subset \mathbb{R}^{3 \times 3 \times N}$.

Rotation invariance and equivariance are two similar notions, and in fact, their relation is more than just resemblance. It is possible to jump from invariance to equivariance and vice versa, just by using the concept of alignment.

**Theorem 1.** *From any mapping $g : \mathbb{R}^{3 \times N} \mapsto \mathbb{R}^D$, a rotation-invariant mapping $g_{inv}$ can be constructed by aligning its input with an equivariant mapping $h_{equi}$.*

*If $h_{equi} : \mathbb{R}^{3 \times N} \mapsto \mathrm{SO}(3)$ is rotation-equivariant*
*And*
$$g_{inv}(x) = g\left(h_{equi}(x)^T x\right) \tag{3}$$

*Then $g_{inv}$ is rotation-invariant.*

**Theorem 2.** *From any rotation-invariant mapping $g_{inv} : \mathbb{R}^{3 \times N} \mapsto \mathbb{R}^{3 \times D}$, a rotation-equivariant mapping $g_{equi}$ can be constructed by aligning its output with an equivariant mapping $h_{equi}$.*

*If $g_{inv} : \mathbb{R}^{3 \times N} \mapsto \mathbb{R}^{3 \times D}$ is rotation-invariant*
*If $h_{equi} : \mathbb{R}^{3 \times N} \mapsto \mathrm{SO}(3)$ is rotation-equivariant*
*And*
$$g_{equi}(x) = h_{equi}(x) g_{inv}(x) \tag{4}$$

*Then $g_{equi}$ is rotation-equivariant.*

Proofs of these two theorems are detailed in the supplementary material, and they are both illustrated in Figure 1.

The first of these two theorems offers a very simple way to turn any operation like a point convolution, into a rotation-invariant mapping. The only requirement is to have the result of a rotation-equivariant mapping, that we call Local Reference Frame (LRF), to perform the alignment. Using the second theorem, we can use the output features of these newly made rotation-invariant operations, to create new LRF that can be used by the following operations. Therefore, we are able to design deep architectures, which update both the rotation-invariant features and their corresponding LRF, from layer to layer.

nated to the input features. Then we go further and **we allow the convolution to use multiple alignments**. We intuitively combine the features obtained by the different alignments in the convolution kernel as explained in Figure 3. As opposed to the previous alignment design, this multi-alignment upgrade has to be defined at the level of the convolution kernel and is therefore specific the type of kernel used. We choose KPConv [27] as our base operator for its great performances and its intuitive way to combine features with a spatially defined kernel.

Eventually, **we design a classification and a segmentation architecture with this multi-alignment scheme, that both perform better than state-of-the-art approaches using rotation-invariant convolutions**. We also provide parameter and ablation studies to attest the performances of each key component of our design.

## 2. Theoretical background

In this section, we define the mathematical framework used in our work.

We use the matrix definition of a point cloud $\mathcal{P} \in \mathbb{R}^{3 \times N}$ and its features $\mathcal{F} \in \mathbb{R}^{D \times N}$. For simplicity, we use the same notation $R \in \mathrm{SO}(3)$ when talking about the linear mapping that is a rotation transformation and its $3 \times 3$ orthogonal matrix representation.

## 3. Related work

We briefly summarize the evolution of 3D deep learning approaches that lead to point convolutional networks and particularly focus on attempts at designing rotation-invariant or rotation-equivariant architectures.

In early stage of 3D deep learning, projective networks were used as they conveniently extend image networks to 3D data. They either project points on multiple 2D images [2, 17, 24], or on volumetric grids [1, 13, 21, 24, 31]. Despite the use of sparse structure [7, 20] and smarter projection grids [23] to alleviate their computational cost, they alter the nature of point clouds and lack flexibility.

Graphs have been proposed as an alternative projection scheme. It is possible to define convolution on graphs [3, 12, 14, 22] or with their spectral representations [5, 35], but the projection of the convolution operator on a surface represented by the graph is not adapted to shape analysis. These graph methods are particularly interesting for topological studies on surfaces, but do not capture volumetric information well. Some recent methods use 3D neighbor graphs [28, 30], which makes them very similar to point cloud methods. They define convolution on graph edge which basically are 3D neighborhoods.

Since the success of PointNet [16, 18], a lot a 3D deep learning methods focused on point clouds directly. The biggest issue in this line of work has been to define a convolution operator compatible to point cloud, efficient, and intuitive. Some works proposed 3D grid-based convolution kernels [9, 10], 2D grid-based tangent convolutions [26], kernels defined by MLPs [8, 11, 29] or by kernel points [27]. These convolution operators are not naturally rotation-invariant, but our alignment method can theoretically be applied to any of them. We choose to use KPConv [27] for its great performances, its intuitive definition with points and its flexibility.

Rotation-invariant deep network have already been proposed in the past. Some works focused on globally invariant architectures, either by aligning the whole input point cloud with Principal Component Analysis (PCA) [32], by using naturally invariant distances and angles point representations [25] or by projecting it to spherical representations [6, 19, 33]. This type of methods is adapted for object models, which have a global orientation, but do not generalize to real scenes where a global orientation is not available, as each particular object has its own orientation. In contrast, our methods can be used on real scenes opening to a lot more applications.

Other locally rotation-invariant methods also exist and are closer to our work. The concurrent work [15] defined an invariant convolution by using spherical harmonics. It was the first paper to stress out the importance of local rotation invariance. Two recent works proposed a rotation-invariant convolution based on the use of distances and angles locally: [37] used a binning strategy making its convolution an oriented 1D operation; [4] used a hierarchical clustering scheme that keeps rotation invariance. Eventually, [36] designed a rotation-invariant kernel with a projection on spherical voxels. All these works struggle to achieve good results on classification and segmentation tasks. Their limitations could be explained by the complexity of their convolution definition, where a standard convolution is simply a linear combination of features in a local neighborhood. Our work is based on standard intuitive point convolutions and we simply use input alignment to create the rotation invariance, which can explain why we outperform all these previous methods in both shape classification and part segmentation.

On the side of these works on rotation-invariant convolutions, but maybe closer to our work, [38] tried to design a rotation-equivariant capsule network. This very innovative work introduced the notion of equivariant LRFs that we also use. However, their alignment scheme is different from ours as it uses a voting scheme and a dynamic rooting for the pose choice. In contrast, we combine multiple alignments directly inside the convolution kernel, effectively combining more information. Furthermore, while they encode their LRFs with quaternions, we choose to keep them as $3 \times 3$ rotation matrices. We outperform their results too, by a large margin.

## 4. Rotation-invariant convolution layer

In this section, we detail the design of our rotation-invariant convolution layer, which is based on KPConv. A standard KPConv layer takes a point cloud $\mathcal{P} \in \mathbb{R}^{3 \times N}$ and its features $\mathcal{F} \in \mathbb{R}^{D \times N}$ as input, and is defined by:

$$g(x, \mathcal{P}, \mathcal{F}) = \sum_{x_i \in \mathcal{N}_x} \left( \sum_{k<K} h\left(x_i - x, \widetilde{x}_k\right) W_k \right) f_i \quad (5)$$

where $x \in \mathbb{R}^3$, $x_i$ are points of $\mathcal{P}$, $f_i$ are their associated features from $\mathcal{F}$, $\widetilde{x}_k$ are the kernel points and $W_k$ are their associated weight matrices. Eventually, we have $\mathcal{N}_x = \left\{ \ x_i \in \mathcal{P} \ \middle| \ \|x_i - x\| \leqslant r \in \mathbb{R} \ \right\}$ the radius neighborhood of $x$; and $h$, a linear influence function defined by:

$$h\left(x_i - x, \widetilde{x}_k\right) = \max\left(0, 1 - \frac{\|(x_i - x) - \widetilde{x}_k\|}{\sigma}\right) \quad (6)$$

We explain how to transform this KPConv definition into a rotation-invariant convolution and how to design deep convolutional architectures with it in the following.

### 4.1. Alignment module

As shown in Figure 2, we design an alignment module to transform a simple KPConv into a rotation-invariant convolution. We first assume that we have a rotation-equivariant

LRF for our input point $x$ and call it $R_x$. This means that using Theorem 1, we could define a rotation-invariant convolution by: $g(R_x^T x, R_x^T \mathcal{P}, \mathcal{F})$.

However, we would only aggregate information from the input features, while we have more. In the context of a convolution layer, we have a rotation-equivariant LRF for each point $x_i \in \mathcal{P}$, that we can call $R_i \in \mathcal{R}$. For lighter notations in the following, we will use $g(x)$ instead of $g(x, \mathcal{P}, \mathcal{F}, \mathcal{R})$. Like the input points, these LRFs can be realigned to become rotation-invariant, and transformed into features with a MLP. Therefore, our aligned convolution is defined by:

$$g_{inv}(x) = \sum_{x_i \in \mathcal{N}_x} \left( \sum_{k<K} h(y_i', \widetilde{x}_k) W_k \right) f_i' \quad (7)$$

or

$$g_{inv}(x) = \sum_{k<K} \left( \sum_{x_i \in \mathcal{N}_x} h(y_i', \widetilde{x}_k) f_i' \right) W_k \quad (8)$$

where $f_i' = \left[ f_i , \text{mlp}(R_x^T R_i) \right]$ and $y_i' = R_x^T (x_i - x)$.

We show in the supplementary material that, as long as $R_x$ is rotation-equivariant, and $\mathcal{F}$ is rotation-invariant, the output of this aligned convolution is rotation-invariant.

We remark that the realigned neighbor LRF $R_x^T R_i$ can be considered as the variation between the neighbor LRF $R_i$ and the center LRF $R_x$. Our alignment module thus provides geometrical information about the neighborhood. We choose to encode this information with an MLP that outputs features of the same dimension as the input features. Therefore, the features provided to the convolution operator are two times larger than in a standard convolution.

### 4.2. Multi-alignment KPConv

Until now, we have assumed having one LRF for each input point, but it is possible to have more than one. To allow this, we design a multi-alignment version of KPConv.

As illustrated in Figure 3, the first step of this multi-alignment convolution consists of using our alignment module with each of the input LRFs, to get different aligned inputs. Note that the MLP encoding the LRF features can be adapted to handle more than one LRF variation per neighbor.

Now instead of naively using a separate convolution kernel for each of these aligned input, we choose to modify the KPConv operations so that the aligned features are merged together. First we apply the kernel correlations to aggregate features of each aligned input on the kernel points. These features, that were previously located at any position in the neighborhood, are now all aggregated on the same fixed locations. It means that we can concatenate them and apply the kernel weights, and effectively aggregate information from all the different alignments.

$$g_{multi}(x) = \sum_{k<K} f_k'' W_k \quad (9)$$

where $f_k''$ is the feature concatenation:

$$f_k'' = \left[ \sum_{x_i \in \mathcal{N}_x} h\left( R_{x,j}^T (x_i - x), \widetilde{x}_k \right) f_i' \right]_{j<J} \quad (10)$$

and $J$ is the number of LRF $R_{x,j}$ associated to $x$

With this multi-alignment modification, the convolution kernel takes features of dimension $J \times 2 \times D$, where $D$ is the dimension of the original input features. To avoid computational overhead, we thus cannot have too many alignments.

### 4.3. LRF initialization and updates

Until now, we have assumed having input LRFs with our input points, but we did not specify how. This is where Theorem 2 becomes handy. The input points of a layer $l$ ($l > 1$) are the query points of the previous layer $l - 1$.

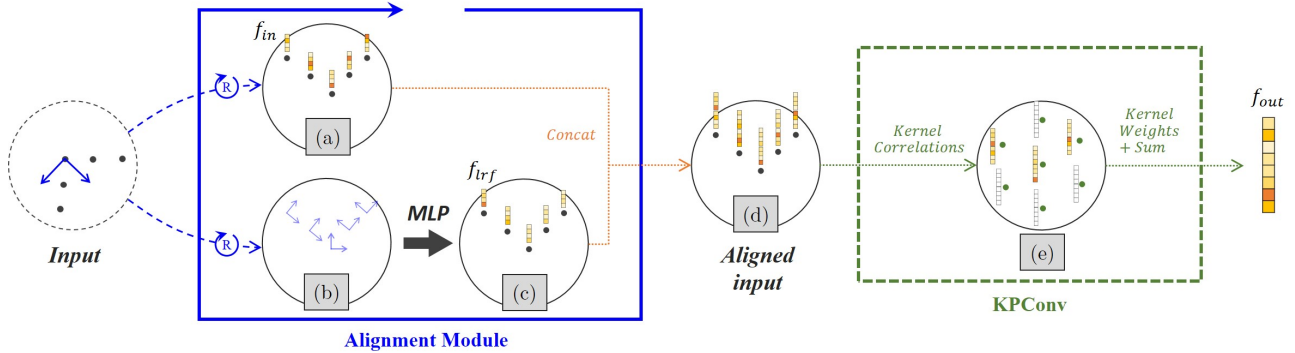

Figure 2. Illustration of our alignment method for point convolution. (a) The input points are aligned with the LRF of the center point. Their corresponding features $f_{in}$ remain the same. (b) The LRFs of the neighbors are also aligned, therefore encoding the LRF variations in this neighborhood. (c) These LRF variations are reduced as rotation-invariant features $f_{lrf}$ by a shared MLP. (d) $f_{in}$ and $f_{lrf}$ are concatenated. (e) The convolution is applied on these aligned points and features.

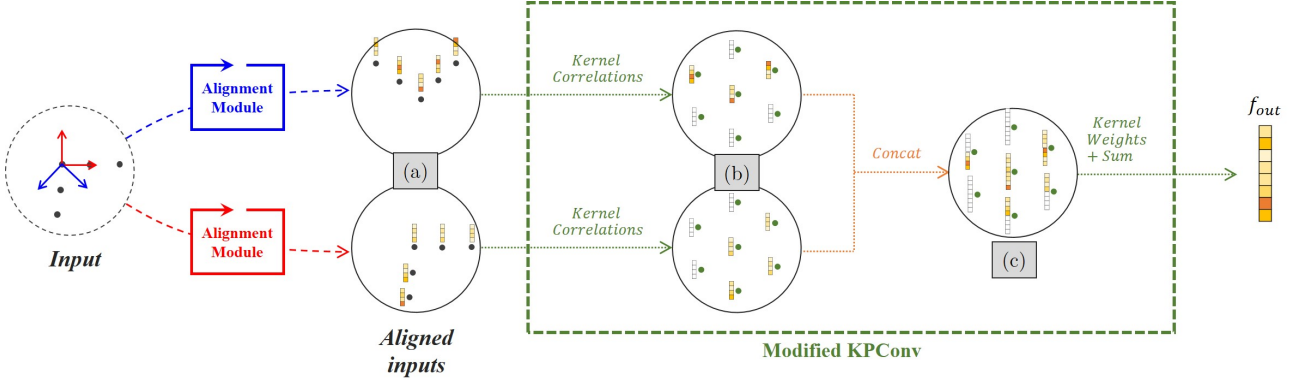

Figure 3. Illustration of our multi-alignment version of KPConv. (a) A set of aligned inputs is obtained by using our alignment module on each input LRF. (b) We apply the first step of KPConv, aggregating features on kernel points via the kernel correlations, for each aligned input. (c) Aggregated features are merged and we apply the KPConv kernel weights to these merged features.

Our convolution outputs rotation-invariant features for every query point. As shown in Figure 4, we simply use a MLP to map these rotation-invariant features to dimension $3 \times 3 \times J$. We thus are in the configuration to use Theorem 2 and transform these features from invariant to equivariant. Each of these rotation-invariant matrices is thus rotated by the corresponding previous LRF. The output of this whole operation becomes rotation-equivariant and can be seen as updated versions of the LRFs we had in the previous layer. Therefore, the network is able to learn new alignments from layer to layer.

A few details still have to be handled. First we need to ensure that these updated LRF are rotation matrices. We follow [16] and use an orthogonalization loss on the invariant MLP output matrices:

$$\mathcal{L}_{ortho} = \omega \sum_{j<J} \left\| I - U_j U_j^T \right\|^2 \quad (11)$$

where $U_j$ are the LRF update matrices predicted by the MLP, and $\omega$ is a parameter controlling the power of orthonormalization.

Then, we also need to handle the "strided" KPConv layers, where LRFs have to be pooled from a point cloud to the next subsampled one. As average is not a well-defined notion in $SO(3)$, we choose to use nearest interpolation: each new point keeps the LRF of its closest neighbor.

Eventually, we also have to initialize the LRFs for the first layer of the network. We choose to use the LRFs obtained from local PCA. They are defined by the eigenvectors of the covariance matrices in local neighborhoods, which by construction are rotation-equivariant. Our network is able to use multiple alignments, which allow us to combine multiple scales when initializing LRFs. In our experiments, we use four scales of k-nearest neighbors: 20, 40, 80 and 160.

### 4.4. Residual network architecture

Like the original KPConv networks [27], we use our rotation-invariant convolutional layer to design architectures with several layers and residual connections. We design two networks, one for object classification, and one for scene/part segmentation.

We use residual connections in our convolution layers, as shown in Figure 4. The black lines describe rotation-invariant features, with the main feature path is on top, where they keep their full dimension. A bottleneck path is derived with MLPs reducing the dimension of the features in the convolution. In blue, we show the rotation-equivariant LRF path, which is decoupled from the rotation-invariant features. The two paths meet when an alignment occurs.

In between the operations shown in Figure 4, we use standard deep learning tools like batch normalization, ReLU, and optional max-pooling and MLPs, in the same way as the original KPConv (See the residual blocks in the supplementary of [27]).

In addition to the convolutional layers, some other layers are used in our architectures. For the classification task, we have a global average pooling, fully connected layers and softmax. For the segmentation architecture, additional

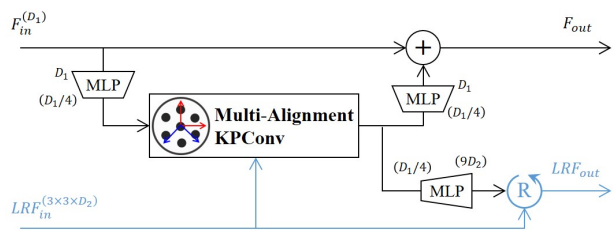

Figure 4. Rotation-invariant residual block with LRF updates. For clarity, we do not show the batch normalization and ReLU blocks, but keep their standard positions (See [27]).

MLPs and nearest upsampling layers are used in the decoder part. All these blocks are rotation-invariant mappings of the features. The only element that could affect rotation invariance is the subsampling operation, that uses a grid which has a certain orientation. Although this has a negligible effect, we ensure that this operation is rotation-equivariant, by using Theorems 1 and 2. We simply align the grid with a global PCA of the input point cloud. We prove the rotation invariance of our layers, and therefore our networks, in the supplementary material.

**Parameters.** We keep the parameters identical to the original KPConv implementation. The same architecture is used with ten convolutional layers, including 4 strided ones. The number of kernel points used in still $K = 15$, and training parameters are not changed. As stated in the last paragraph, we use $J = 4$ LRFs per point. The orthonormalization loss power $\omega = 0.5$ is chosen in the following experiments (See Section 5.4 for a study of this parameter).

## 5. Experimental results

In this section, we report our results on three tasks: object classification, object part segmentation and scene segmentation. Furthermore, we provide parameters and ablation studies. Our implementation is based on the newer PyTorch implementation provided by KPConv authors, and will be published upon publication.

We train our networks on a DGX station with an Intel(R) Xeon(R) E5-2698 v4 CPU and four Tesla V100 DGXS GPU, using only one GPU at a time. Our training times are slower than the original implementation due to the use of multiple alignments and the computational overhead that it implies. For the classification task, it takes about 5 hours to do 100 training epochs, while it takes less than 3 hours for the part segmentation task. The scene segmentation is a bit slower, with 6 hours for 100 epochs. In all the tasks, the networks converge after 150 or 200 epochs.

### 5.1. Object classification

First our classification network is evaluated on ModelNet40 dataset [31]. It consists of $12,311$ meshed CAD models from 40 categories of common objects. We used the preprocessed point clouds from [18], which are separated into a training set of d $9,843$ models and a test set of $2,468$ models. The input point clouds are rescaled to fit in a unit sphere and subsampled by a grid with 2cm cells like in [27]. The number of points per model is thus variable, and its average is the same as in the original KPConv implementation. We also use the same constant 1 input feature which is obviously rotation-invariant.

Following a common practice [15, 33, 36], we refer to a network trained/tested with random rotation augmentation in SO(3) as A for "Augmentation", and to a network trained/tested without augmentation as N for "No augmentation". We compare our method to concurrent works in two scenarios: N/A to test rotation invariance and A/A to compare with methods that are not strictly invariant. Some method do not use the N/A scenario but a commonly called $z$/SO(3) scenario where the network was trained with rotations around the vertical axis. For benchmark purpose we provide our result is this scenario too, as it does not make any difference for a rotation-invariant network.

| Methods | N/A | $z$/SO(3) | A/A |
|---|---|---|---|
| PRIN [36] | 70.4 | - | 80.1 |
| SphCNN [6] | - | 78.6 | 86.9 |
| GICOPix [33] | 84.0 | - | 86.3 |
| SPHNet [15] | 86.6 | - | 87.6 |
| KPConv [27] | 38.9 | - | 87.7 |
| QuatCaps [38] | 74.1 | - | - |
| RICNN [37] | - | 86.4 | 86.4 |
| SRINet [25] | 87.0 | - | 87.0 |
| ClusterNet [4] | - | 87.1 | 87.1 |
| MA-KPC *(Ours)* | **89.1** | **89.1** | **89.1** |

Table 1. Comparison of robust and invariant methods to rotation on ModelNet40. Overall Accuracy is compared in different challenging scenarios.

Table 1 shows that our method outperforms all concurrent approaches, whether they are strictly rotation-invariant or just robust to rotations, in any scenario. These results also confirm the strict rotation invariance of our network, as we get the same score in all the scenarios.

### 5.2. Object Part Segmentation

Our segmentation architecture is then evaluated on the ShapeNetPart dataset [34]. This dataset is an object part segmentation task, which contains $16,681$ point clouds from 16 categories, each with 2-6 part labels. The goal is to give the label of the part each points belongs too, knowing which type of object it is from.

Here again, we use data provided by [18], and follow the procedure of [27]: standard train/test splits, input clouds rescaled to unit sphere and 2cm subsampling. The only dif-ference here is in the input features. As we design a method that should be invariant to rotation augmentation, we do not use the original $x$, $y$, $z$ coordinates as input features. It would be cheating as it would give the information about the object orientation to the network. Therefore, like for the object classification task, we only use a constant 1 feature as input.

| Methods | N/A | $z$/SO(3) | A/A |
|---|---|---|---|
| PRIN [36] | 54.2 | - | - |
| RICNN [37] | - | 75.3 | 75.5 |
| SRINet [25] | 76.9 | - | 76.9 |
| KPConv [27] | 54.5 | - | 77.6 |
| MA-KPC *(Ours)* | **79.5** | **79.5** | **79.5** |

Table 2. Comparison of robust and invariant methods to rotation on ShapeNetPart. We compare the mean class IoU metric in different challenging scenarios.

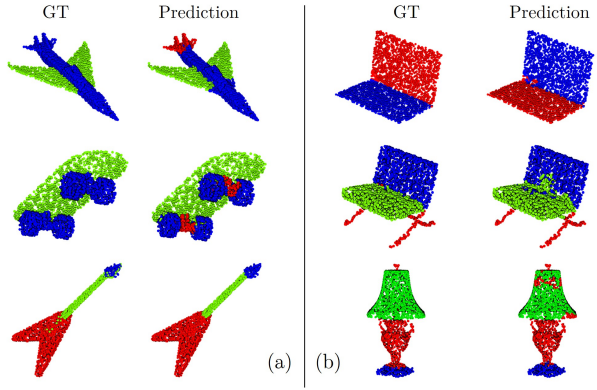

Figure 5. Examples of ShapetPart segmentation results. On the left (a), we show good segmentations where our network even produced better parts than the groundtruth. On the right (b) we show failure cases and errors.

Like for object classification, we show the results in the three scenarios N/A, $z$/SO(3) and A/A. We measure the performances with the mean class IoU metric, which is defined as the average of all class IoU. In Table 2, we see that, on this task too, our network outperforms concurrent rotation-invariant networks in all scenarios.

We provide detailed class IoU for our method in Table 3. In Figure 5 we show examples of shapes segmented with our method. We can see that our method is able to detect object parts very well in most cases, sometimes even correcting annotation errors. However, we still see some failure cases, especially in case where orientation is not given away by the shape like for laptops. Despite outperforming state-of-the-art rotation-invariant approaches, our method still has limitations, which we develop more in the conclusion.

### 5.3. Ablation study

In this section we study the effect of removing three key components of our networks: the multiple alignments, the LRF feature merging, and the locality of the rotation invariance. We compare the full version of our approach to three ablated versions: the *No-merge* version, without the LRF features $f_{lrf}$ normally merged with $f_{in}$ in the convolution; the *1-local* version, with only one LRF per point in the convolution layers (initialized with PCA on $k = 20$ neighborhoods); and the *1-global*, also with one LRF per point, but initialized with a global PCA instead of a local one. This last version is still globally invariant, but the convolutions are not locally invariant anymore.

We keep the same experimental conditions and compare the different versions on ModelNet40 (Table 4) and ShapeNetPart (Table 3). We can see that for both classification and segmentation tasks, all the key components are essential to achieve the best performances.

| Version  | 1-global | 1-local | No-merge | Full |
|----------|----------|---------|----------|------|
| Accuracy | 87.3     | 88.5    | 88.7     | 89.1 |

Table 4. Ablation study on ModelNet40. Overall Accuracy is compared for different numbers and types of LRFs.

### 5.4. Influence of orthonormalization loss

Here, we study the effect that orthonormalization loss has on our network convergence and performances. For this we evaluate the performances of our network on ModelNet40 object classification task, for different values of the parameter $\omega$. Figure 6 shows the results of this study. We can see that this parameter has an optimal value around 0.5.

Although the effect of $\omega$ could seem obscure as one of many parameters in this big architecture, it can be explained quite intuitively. Think of $\omega$ as the importance of the orthonormalization loss compared to the output loss. If this loss value is ten times bigger than the output loss, then the gradient of the learnable parameters are going to be ten times more influenced by this loss than by the output loss. As these two losses rapidly decrease during training, they will reach an equilibrium and keep it until convergence. A low $\omega$ means the orthonormalization loss will reach its lowest value later, letting more freedom to the LRF updates which could lead to aberrations. A high $\omega$ value means the orthonormalization loss will decrease faster, with less freedom for the network to learn interesting LRFs. This is why we see a concave plot in Figure 6, with an ideal value of $\omega = 0.5$.

| Version  | class avg. | inst. avg. | aero | bag  | cap  | car  | chair | ear  | guit | knif | lamp | lapt | moto | mug  | pist | rock | skate | table |
|----------|------------|------------|------|------|------|------|-------|------|------|------|------|------|------|------|------|------|-------|-------|
| 1-global | 78.1       | 79.7       | 79.4 | 79.9 | 83.8 | 73.2 | 84.1  | 76.6 | 91.4 | 83.9 | 76.3 | 79.5 | 64.2 | 91.3 | 79.7 | 51.3 | 78.7  | 76.8  |
| 1-local  | 78.9       | 80.0       | 80.6 | 78.7 | 84.3 | 73.5 | 84.8  | 79.3 | 90.8 | 83.7 | 76.7 | 74.7 | 70.2 | 94.6 | 77.0 | 57.8 | 79.4  | 76.7  |
| No-merge | 79.3       | 80.5       | 81.3 | 80.8 | 84.6 | 74.1 | 85.5  | 79.1 | 90.9 | 84.8 | 75.6 | 76.7 | 70.9 | 94.5 | 76.8 | 55.8 | 80.6  | 77.3  |
| Full     | 79.5       | 80.9       | 80.8 | 78.8 | 84.8 | 75.8 | 86.3  | 79.9 | 91.0 | 82.2 | 76.2 | 77.2 | 70.4 | 94.9 | 79.5 | 56.1 | 80.1  | 77.7  |

Table 3. Ablation study on ShapeNetPart (N/A scenario). We provide all class IoU for different numbers and types of LRFs.

| Method | mIoU | ceil. | floor | wall | beam | col. | wind. | door | chair | table | book. | sofa | board | clut. |
|---|---|---|---|---|---|---|---|---|---|---|---|---|---|---|
| Pointnet [16] | 41.1 | 88.8 | 97.3 | 69.8 | 0.1 | 3.9 | 46.3 | 10.8 | 52.6 | 58.9 | 40.3 | 5.9 | 26.4 | 33.2 |
| TangentConv [26] | 52.6 | 90.5 | 97.7 | 74.0 | 0.0 | **20.7** | 39.0 | 31.3 | 69.4 | 77.5 | 38.5 | 57.3 | 48.8 | 39.8 |
| ParamConv [29] | 58.3 | 92.3 | 96.2 | 75.9 | **0.3** | 6.0 | **69.5** | 63.5 | 66.9 | 65.6 | 47.3 | **68.9** | 59.1 | 46.2 |
| KPConv [27] | **65.4** | **92.6** | 97.3 | **81.4** | 0.0 | 16.5 | 54.5 | **69.5** | 90.1 | **80.2** | **74.6** | 66.4 | **63.7** | **58.1** |
| MA-KPC *(Ours)* | 62.1 | 91.9 | **97.9** | 80.0 | 0.0 | 14.4 | 55.2 | 65.3 | **90.6** | 78.0 | 71.2 | 59.3 | 51.5 | 51.9 |

Table 5. Semantic segmentation IoU scores on S3DIS *Area-5*. Comparison with non-invariant methods in z/z scenario

## 5.5. Scene segmentation

Our last experiment evaluates our segmentation architecture on the S3DIS indoor scene segmentation dataset. It contains six large-scale indoor areas from three different buildings for a total of 273 million points annotated with 13 classes. We use the same experimentation procedure as in [27], and use Area-5 as test scene. In accordance to the rest of our experiments, we use the same parameters as the original KPConv paper. To allow our method to have a sense of the vertical direction, like the other standard methods do, we use the height coordinate as input feature, with the RGB colors and a constant 1 for geometry.

To our knowledge, we are the first to present results of a rotation-invariant deep network on scene segmentation. The reason is that, in this type of dataset, it is very common to have shapes naturally aligned by gravity. We could test our method on randomly oriented scenes, but this would not have any practical application as this natural vertical direction is nearly always known. Therefore, the possible rotation augmentation is reduced to one single axis, simplifying the problem for standard non-invariant methods, and the advantage of using a rotation-invariant method is limited. However, we believe that in some cases, a rotation-invariant network has some advantages, for example to segment complex shapes likely to be rotated a lot.

The results presented in Table 5 show that our method performs very well on this task, only conceding to the original KPConv, which is top of the state of the art on this dataset. We can see that on such a dataset, the gap between rotation-invariant and standard deep networks is very small with only 3 MIoU points. The *chair* IoU provides a very interesting insight on the method. We see that our method performs particularly well on this class. In this dataset, the *chair* class is probably the class with a consistent shape that has the most random orientations with respect to walls. This confirms our intuition that a rotation-invariant method should perform better on complex shapes that are likely to be rotated a lot.

## 6. Conclusion

In this work, we studied how rotation invariance and rotation equivariance are related, and used this relation to design a rotation-invariant convolutional architecture. We presented an alignment method that can introduce rotation invariance into any convolution operator. It uses equivariant LRFs to locally align convolution input points, and it combines neighbor LRFs with input features. Our design, using the flexible KPConv operator, is able to use multiple alignments for better results. With this core convolutional layer, we built classification and segmentation architectures that proved to be superior to earlier rotation-invariant methods and reduced the performance gap with standard deep networks.

The fact that our rotation-invariant architecture does not reach the performances of its non-invariant counterpart means that there is still a lot to explore in this field. We believe one of the limitations of our method is that using invariant representations reduces the diversity of shapes that a convolution has to learn, which can harm the ability of the network to generalize to new shapes. Further work could explore this problem. We also need to better understand how the network is able to learn jointly the LRF alignments and the convolution kernel weights. The latter depends on the former and vice versa. Eventually, we can imagine the network learning its own alignment from scratch, instead of using PCA as input.

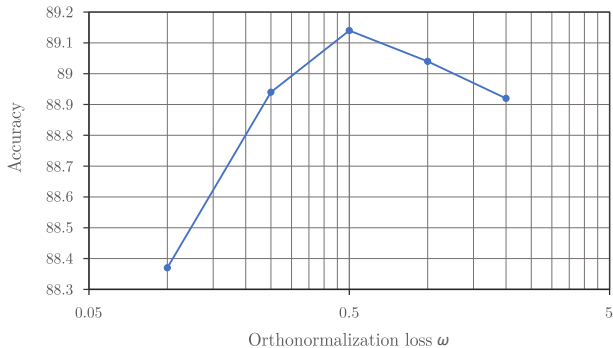

Figure 6. Influence of orthonormalization loss parameters $\omega$ on ModelNet40 classification accuracy.

# References

[1] Y. Ben-Shabat, M. Lindenbaum, and A. Fischer. 3dmfv: Three-dimensional point cloud classification in real-time using convolutional neural networks. *IEEE Robotics and Automation Letters*, 3(4):3145–3152, 2018. 3

[2] A. Boulch, B. Le Saux, and N. Audebert. Unstructured point cloud semantic labeling using deep segmentation networks. In *Proceedings of the Workshop on 3D Object Retrieval (3DOR)*, 2017. 3

[3] M. M. Bronstein, J. Bruna, Y. LeCun, A. Szlam, and P. Vandergheynst. Geometric deep learning: going beyond euclidean data. *IEEE Signal Processing Magazine*, 34(4):18–42, 2017. 3

[4] C. Chen, G. Li, R. Xu, T. Chen, M. Wang, and L. Lin. Clusternet: Deep hierarchical cluster network with rigorously rotation-invariant representation for point cloud analysis. In *Proceedings of the IEEE Conference on Computer Vision and Pattern Recognition*, pages 4994–5002, 2019. 1, 3, 6

[5] M. Defferrard, X. Bresson, and P. Vandergheynst. Convolutional neural networks on graphs with fast localized spectral filtering. In *Advances in Neural Information Processing Systems*, pages 3844–3852, 2016. 3

[6] C. Esteves, C. Allen-Blanchette, A. Makadia, and K. Daniilidis. Learning so (3) equivariant representations with spherical cnns. In *Proceedings of the European Conference on Computer Vision (ECCV)*, pages 52–68, 2018. 1, 3, 6

[7] B. Graham, M. Engelcke, and L. van der Maaten. 3d semantic segmentation with submanifold sparse convolutional networks. In *Proceedings of the IEEE Conference on Computer Vision and Pattern Recognition*, pages 9224–9232, 2018. 3

[8] P. Hermosilla, T. Ristchel, P.-P. Vázquez, Á. Vinacua, and T. Ropinski. Monte carlo convolution for learning on non-uniformly sampled point clouds. *ACM Transactions on Graphics (TOG)*, 37(6):235–1, 2018. 3

[9] B.-S. Hua, M.-K. Tran, and S.-K. Yeung. Pointwise convolutional neural networks. In *Proceedings of the IEEE Conference on Computer Vision and Pattern Recognition*, pages 984–993, 2018. 3

[10] T. Le and Y. Duan. Pointgrid: A deep network for 3d shape understanding. In *Proceedings of the IEEE Conference on Computer Vision and Pattern Recognition*, pages 9204–9214, 2018. 3

[11] Y. Li, R. Bu, M. Sun, W. Wu, X. Di, and B. Chen. Pointcnn: Convolution on x-transformed points. In *Advances in Neural Information Processing Systems*, pages 820–830, 2018. 3

[12] J. Masci, D. Boscaini, M. Bronstein, and P. Vandergheynst. Geodesic convolutional neural networks on riemannian manifolds. In *Proceedings of the IEEE international conference on computer vision workshops*, pages 37–45, 2015. 3

[13] D. Maturana and S. Scherer. Voxnet: A 3d convolutional neural network for real-time object recognition. In *Intelligent Robots and Systems (IROS), 2015 IEEE/RSJ International Conference on*, pages 922–928. IEEE, 2015. 3

[14] F. Monti, D. Boscaini, J. Masci, E. Rodola, J. Svoboda, and M. M. Bronstein. Geometric deep learning on graphs and manifolds using mixture model cnns. In *Proceedings of the IEEE Conference on Computer Vision and Pattern Recognition*, pages 5115–5124, 2017. 3

[15] A. Poulenard, M.-J. Rakotosaona, Y. Ponty, and M. Ovsjanikov. Effective rotation-invariant point cnn with spherical harmonics kernels. In *2019 International Conference on 3D Vision (3DV)*, pages 47–56. IEEE, 2019. 1, 3, 6

[16] C. R. Qi, H. Su, K. Mo, and L. J. Guibas. Pointnet: Deep learning on point sets for 3d classification and segmentation. In *Proceedings of the IEEE Conference on Computer Vision and Pattern Recognition*, pages 652–660, 2017. 3, 5, 8

[17] C. R. Qi, H. Su, M. Nießner, A. Dai, M. Yan, and L. J. Guibas. Volumetric and multi-view cnns for object classification on 3d data. In *Proceedings of the IEEE conference on Computer Vision and Pattern Recognition*, pages 5648–5656, 2016. 3

[18] C. R. Qi, L. Yi, H. Su, and L. J. Guibas. Pointnet++: Deep hierarchical feature learning on point sets in a metric space. In *Advances in Neural Information Processing Systems*, pages 5099–5108, 2017. 3, 6

[19] Y. Rao, J. Lu, and J. Zhou. Spherical fractal convolutional neural networks for point cloud recognition. In *Proceedings of the IEEE Conference on Computer Vision and Pattern Recognition*, pages 452–460, 2019. 1, 3

[20] G. Riegler, A. O. Ulusoy, and A. Geiger. Octnet: Learning deep 3d representations at high resolutions. In *Proceedings of the IEEE Conference on Computer Vision and Pattern Recognition*, volume 3, 2017. 3

[21] X. Roynard, J.-E. Deschaud, and F. Goulette. Classification of point cloud scenes with multiscale voxel deep network. *arXiv preprint arXiv:1804.03583*, 2018. 3

[22] M. Simonovsky and N. Komodakis. Dynamic edge-conditioned filters in convolutional neural networks on graphs. In *Proceedings of the IEEE conference on computer vision and pattern recognition*, pages 3693–3702, 2017. 3

[23] H. Su, V. Jampani, D. Sun, S. Maji, E. Kalogerakis, M.-H. Yang, and J. Kautz. Splatnet: Sparse lattice networks for point cloud processing. In *Proceedings of the IEEE Conference on Computer Vision and Pattern Recognition*, pages 2530–2539, 2018. 3

[24] H. Su, S. Maji, E. Kalogerakis, and E. Learned-Miller. Multi-view convolutional neural networks for 3d shape recognition. In *Proceedings of the IEEE International Conference on Computer Vision*, pages 945–953, 2015. 3

[25] X. Sun, Z. Lian, and J. Xiao. Srinet: Learning strictly rotation-invariant representations for point cloud classification and segmentation. In *Proceedings of the 27th ACM International Conference on Multimedia*, pages 980–988, 2019. 1, 3, 6

[26] M. Tatarchenko, J. Park, V. Koltun, and Q.-Y. Zhou. Tangent convolutions for dense prediction in 3d. In *Proceedings of the IEEE Conference on Computer Vision and Pattern Recognition*, pages 3887–3896, 2018. 3, 8

[27] H. Thomas, C. R. Qi, J.-E. Deschaud, B. Marcotegui, F. Goulette, and L. J. Guibas. Kpconv: Flexible and deformable convolution for point clouds. In *Proceedings of the IEEE International Conference on Computer Vision*, pages 6411–6420, 2019. 2, 3, 5, 6, 8

[28] N. Verma, E. Boyer, and J. Verbeek. Feastnet: Feature-steered graph convolutions for 3d shape analysis. In *Proceedings of the IEEE Conference on Computer Vision and Pattern Recognition*, pages 2598–2606, 2018. 3

[29] S. Wang, S. Suo, W.-C. Ma, A. Pokrovsky, and R. Urtasun. Deep parametric continuous convolutional neural networks. In *Proceedings of the IEEE Conference on Computer Vision and Pattern Recognition*, pages 2589–2597, 2018. 3, 8

[30] Y. Wang, Y. Sun, Z. Liu, S. E. Sarma, M. M. Bronstein, and J. M. Solomon. Dynamic graph cnn for learning on point clouds. *ACM Transactions on Graphics (TOG)*, 38(5):1–12, 2019. 3

[31] Z. Wu, S. Song, A. Khosla, F. Yu, L. Zhang, X. Tang, and J. Xiao. 3d shapenets: A deep representation for volumetric shapes. In *Proceedings of the IEEE conference on Computer Vision and Pattern Recognition*, pages 1912–1920, 2015. 3, 6

[32] Z. Xiao, H. Lin, R. Li, L. Geng, H. Chao, and S. Ding. Endowing deep 3d models with rotation invariance based on principal component analysis. In *2020 IEEE International Conference on Multimedia and Expo (ICME)*, pages 1–6. IEEE, 2020. 1, 3

[33] Q. Yang, C. Li, W. Dai, J. Zou, G.-J. Qi, and H. Xiong. Rotation equivariant graph convolutional network for spherical image classification. In *Proceedings of the IEEE Conference on Computer Vision and Pattern Recognition*, pages 4303–4312, 2020. 1, 3, 6

[34] L. Yi, V. G. Kim, D. Ceylan, I. Shen, M. Yan, H. Su, C. Lu, Q. Huang, A. Sheffer, L. J. Guibas, et al. A scalable active framework for region annotation in 3d shape collections. *ACM Transactions on Graphics (TOG)*, 35(6):210, 2016. 6

[35] L. Yi, H. Su, X. Guo, and L. J. Guibas. Syncspeccnn: Synchronized spectral cnn for 3d shape segmentation. In *CVPR*, pages 6584–6592, 2017. 3

[36] Y. You, Y. Lou, Q. Liu, Y.-W. Tai, L. Ma, C. Lu, and W. Wang. Pointwise rotation-invariant network with adaptive sampling and 3d spherical voxel convolution. In *AAAI*, pages 12717–12724, 2020. 1, 3, 6

[37] Z. Zhang, B.-S. Hua, D. W. Rosen, and S.-K. Yeung. Rotation invariant convolutions for 3d point clouds deep learning. In *2019 International Conference on 3D Vision (3DV)*, pages 204–213. IEEE, 2019. 1, 3, 6

[38] Y. Zhao, T. Birdal, J. E. Lenssen, E. Menegatti, L. Guibas, and F. Tombari. Quaternion equivariant capsule networks for 3d point clouds. In *Proceedings of the European Conference on Computer Vision (ECCV)*, 2020. 1, 3, 6

# *Supplementary Material for*
# Rotation-Invariant Point Convolution With Multiple Equivariant Alignments.

## A. Theorem proofs

**Theorem 1:** Proof.

Let $g$ be a mapping from $\mathbb{R}^{3 \times N}$ to $\mathbb{R}^D$.

Let $h_{equi} : \mathbb{R}^{3 \times N} \mapsto \mathrm{SO}(3)$ be a rotation-equivariant mapping.

And let us define the mapping $g_{inv} : \mathbb{R}^{3 \times N} \mapsto \mathbb{R}^D$ by:
$$g_{inv}(x) = g\left(h_{equi}(x)^T x\right)$$

Let $x \in \mathbb{R}^{3 \times N}$ be a real matrix and $R \in \mathrm{SO}(3)$ be a rotation matrix. Then:
$$g_{inv}(Rx) = g\left(h_{equi}(Rx)^T Rx\right)$$

With $h_{equi}$ equivariance:
$$g_{inv}(Rx) = g\left((R\, h_{equi}(x))^T Rx\right)$$
$$g_{inv}(Rx) = g\left(h_{equi}(x)^T R^T Rx\right)$$

As $R$ is a rotation $R^T R = I$. Therefore:
$$g_{inv}(Rx) = g\left(h_{equi}(x)^T x\right)$$

We have thus proved that $\forall x \in \mathbb{R}^{3 \times N}$ and $\forall R \in \mathrm{SO}(3)$
$$g_{inv}(Rx) = g_{inv}(x)$$

**Theorem 2:** Proof.

Let $g_{inv} : \mathbb{R}^{3 \times N} \mapsto \mathbb{R}^{3 \times D}$ be a rotation-invariant mapping.

Let $h_{equi} : \mathbb{R}^{3 \times N} \mapsto \mathrm{SO}(3)$ be a rotation-equivariant mapping.

And let us define the mapping $g_{equi} : \mathbb{R}^{3 \times N} \mapsto \mathbb{R}^{3 \times D}$ by:
$$g_{equi}(x) = h_{equi}(x) g_{inv}(x)$$

Let $x \in \mathbb{R}^{3 \times N}$ be a real matrix and $R \in \mathrm{SO}(3)$ be a rotation matrix. Then:
$$g_{equi}(Rx) = h_{equi}(Rx) g_{inv}(Rx)$$

With $h_{equi}$ equivariance and $g_{inv}$ invariance:
$$g_{equi}(Rx) = R\, h_{equi}(x) g_{inv}(x)$$

We have thus proved that $\forall x \in \mathbb{R}^{3 \times N}$ and $\forall R \in \mathrm{SO}(3)$
$$g_{equi}(Rx) = R\, g_{equi}(x)$$

The proofs of these two theorems are straightforward. It is more tricky to understand how we use them to prove that our network is rotation-invariant.

## B. Network invariance proof

To prove our network invariance, we are going to prove that every successive layer is rotation-invariant. A network layer is a mapping function that takes several inputs: The input points $\mathcal{P} \in \mathbb{R}^{3 \times N}$, the input features $\mathcal{F} \in \mathbb{R}^{D \times N}$, the input LRFs $\mathcal{R} \in \mathrm{SO}(3)^N$, the query points $\mathcal{X} \in \mathbb{R}^{3 \times N'}$, and the query LRFs $\mathcal{R}' \in \mathrm{SO}(3)^{N'}$

The query point cloud $\mathcal{X}$ can be the same as the input point cloud or can be a subsampled version of it. As explained in the main paper, our subsampling operation is rotation-invariant, which is a necessary condition for the next proofs.

All these inputs do not react the same to a rotation. In the following we always consider that the input features are rotation-invariant, and that all the other inputs are rotation-equivariant, including the query points (which is why we need a rotation-equivariant subsampling). Therefore, when we consider the input of a layer to be rotated by a rotation $R \in \mathrm{SO}(3)$, we have:

$$R(\mathcal{X}, \mathcal{R}', \mathcal{P}, \mathcal{F}, \mathcal{R}) = (R\mathcal{X}, R\mathcal{R}', R\mathcal{P}, \mathcal{F}, R\mathcal{R})$$

In the following and in the main paper we use a single query point $x \in \mathbb{R}^3$ for clarity, and its associated LRF $R_x \in$

SO(3) but every proof here generalizes to a set of query points $\mathcal{X}$ and their associated LRFs $\mathcal{R}'$.

Lets first prove the rotation-invariance of our convolution layer, in its full version, which is the most general case.

**Convolution layer rotation-invariance:** Proof.

We use the notations introduced in the main paper and use our convolution layer definition:

$$g(x, \mathcal{P}, \mathcal{F}, \mathcal{R}) = \sum_{k<K} f''_k W_k$$

Where:

$$f''_k(x, \mathcal{P}, \mathcal{F}, \mathcal{R}) = \left[ \sum_{x_i \in \mathcal{N}_x} h\left(R^T_{x,j}(x_i - x), \widetilde{x}_k\right) f'_i(x_i) \right]_{j<J}$$

And:

$$f'_i(x_i, x, \mathcal{P}, \mathcal{F}, \mathcal{R}) = \left[ f_i(x_i), \text{mlp}(R^T_x R_i) \right]$$

We can thus prove the rotation-invariance step by step. Let us first prove that $f'_i$ is rotation-invariant. First we know that $f_i$ is rotation invariant because it is a feature from $\mathcal{F}$. Then we know that $R_x$ is rotation-equivariant because it is a LRF from $\mathcal{R}'$, therefore with Theorem 1, we know that $\text{mlp}(R^T_x R_i)$ is rotation-invariant.

As a concatenation of two rotation-invariant features, $f'_i$ is rotation-invariant.

Now we need to prove that $f''_k$ is rotation-invariant, which we can do for each $j < J$ independently. Let us call

$$f''_{k,j}(x, \mathcal{P}, \mathcal{F}, \mathcal{R}) = \sum_{x_i \in \mathcal{N}_x} h\left(R^T_{x,j}(x_i - x), \widetilde{x}_k\right) f'_i(x_i)$$

We know that $R^T_{x,j}$ is rotation-equivariant, and we also know that. And as $\forall R \in \text{SO}(3)$:

$$(Rx_i - Rx) = R(x_i - x)$$

Therefore, using Theorem 1, $h\left(R^T_{x,j}(x_i - x), \widetilde{x}_k\right)$ is rotation-invariant.

The remaining operation in $f''_{k,j}$ is the summation over $\mathcal{N}_x$. And as $\mathcal{N}_x$ is a radius neighborhood, we know this sum is rotation-invariant too. As a consequence, we have proven the rotation invariance of $f''_{k,j}$ and therefore $f''_k$.

The last step of the proof is very straightforward, with $f''_k$ being rotation-invariant, and $K$ and $W_k$ being fixed parameters, g is rotation-invariant too.

Q.E.D.

**Other layers rotation-invariance:** Proof.

Most of the other layers of the network are just feature mappings, which means they only take the features as input:

$$g(x, R_x, \mathcal{P}, \mathcal{F}, \mathcal{R}) = g(\mathcal{F})$$

This automatically makes them rotation-invariant. It is the case for the MLPs, ReLU, batch normalization, global average pooling, fully connected layers and softmax blocks. The only operation left are the the max-pooling used on residual shortcuts, and the upsampling in the decoder part of our segmentation architecture. Same as in the neighbors in the convolution, the max-pooling is computed over spherical neighborhoods, and is thus rotation-invariant. Eventually the nearest neighbors upsampling is also based on radius, and thus invariant by rotation.

As a consequence, the input and output features of each layer of the network are rotation-invariant. Making the whole network rotation-invariant

**Note on invariance locality.**

These proofs concerned the global invariance of our network. Here we discuss about the level of local invariance our network have.

Following the same reasoning as before, we can prove that our convolution operation is locally invariant. At a certain point, if you rotate the neighbors and there LRF in the spherical neighborhood, the convolution will output the same values. However, this only holds if the features are the same too, and if you look at two parts of a point cloud with identical local shapes modulo a rotation, the features are not necessarily the same, because of their receptive field.

Local invariance is thus tightly related to the notion of receptive fields. As a reminder, the receptive field is defined for a single point at a certain layer of the network, and consists of all the points of the input that have an influence on the features of this particular point. For example, after the first convolution layer, the receptive field at a given point is its spherical neighborhood, and the receptive field grows in size as the network grows in depth.

For the features of the convolution neighbors to be the same, you have to rotate the whole receptive field, and not only the spherical neighborhood. Therefore, the convolution layers, which have a larger and larger receptive field, will become more and more globally invariant instead of locally invariant.

It is important to understand that strict local invariance is not desirable in the network. We want the features to have a large receptive field and depend from the shape of their surroundings. For example the corner of a chair and a table can have the same shape locally, but we need information from larger surroundings to differentiate it. Having this in mind, we encourage the network to build complex shape representations as soon as in the first layer, by feeding it LRFs from different scales.